%% file: manuscript.tex
\documentclass{article}
\usepackage{spconf,amsmath,graphicx}

\usepackage{graphicx}
\usepackage{amsmath}
\usepackage{amssymb}
\usepackage{booktabs}

\usepackage{xcolor} 
\usepackage{times}
\usepackage{epsfig}
\usepackage{graphicx}
\usepackage{amsmath}
\usepackage{amssymb}
\usepackage[bb=dsserif]{mathalpha}
\usepackage{bm}

\usepackage{booktabs,colortbl,tabularx}
\usepackage{pifont}%

\usepackage{mathtools}
\usepackage{commath}
\usepackage{algorithm}
\usepackage{algpseudocode}

\usepackage{multirow}
\usepackage{comment} 
\usepackage{enumitem}

\definecolor{Gray}{gray}{0.9}

\newcommand{\cmark}{\ding{51}}%
\newcommand{\xmark}{\ding{55}}%

\definecolor{battleshipgrey}{rgb}{0.52, 0.52, 0.51}

\title{Tree of Uncertain Thoughts Reasoning for Large Language Models}
\name{Shentong Mo$^{1,2}$ \qquad Miao Xin$^{3}$\sthanks{Corresponding author.}}
\address{$^1$Carnegie Mellon University, $^2$MBZUAI, $^3$Institute of Automation, Chinese Academy of Sciences}

\begin{document}
%
\maketitle
\input{SECTIONS/00_Abstract/manuscript}

\input{SECTIONS/10_Introduction/manuscript}

\input{SECTIONS/20_Related_Work/manuscript}

\input{SECTIONS/30_Method/manuscript}

\input{SECTIONS/40_Experiments/manuscript}

\input{SECTIONS/50_Ablation_Study/manuscript}

\input{SECTIONS/60_Conclusion/manuscript}

\vfill\pagebreak

\bibliographystyle{IEEEbib}
\bibliography{reference}

\end{document}

%% file: SECTIONS/00_Abstract/manuscript.tex
\begin{abstract}

    While the recently introduced Tree of Thoughts (ToT) has heralded advancements in allowing Large Language Models (LLMs) to reason through foresight and backtracking for global decision-making, it has overlooked the inherent local uncertainties in intermediate decision points or ``thoughts". 
    These local uncertainties, intrinsic to LLMs given their potential for diverse responses, remain a significant concern in the reasoning process. 
    Addressing this pivotal gap, we introduce the Tree of Uncertain Thoughts (TouT) -- a reasoning framework tailored for LLMs.
    Our TouT effectively leverages Monte Carlo Dropout to quantify uncertainty scores associated with LLMs' diverse local responses at these intermediate steps.
    By marrying this local uncertainty quantification with global search algorithms, TouT enhances the model's precision in response generation. 
    We substantiate our approach with rigorous experiments on two demanding planning tasks: Game of 24 and Mini Crosswords. 
    The empirical evidence underscores TouT's superiority over both ToT and chain-of-thought prompting methods.
    
\end{abstract}

%% file: SECTIONS/10_Introduction/manuscript.tex
\section{Introduction}

Modern Large-scale Language Models (LLMs), including GPT's early iterations~\cite{Radford2018gpt,Radford2019language,brown2020language}, the recent GPT-4~\cite{openai2023gpt4}, and LLaMA-2~\cite{touvron2023llama}, have showcased remarkable prowess in tasks that demand mathematical, symbolic, commonsense, and knowledge reasoning. 
Despite this, their reasoning process primarily hinges on the autoregressive mechanism, sequentially generating text and making token-level decisions from left-to-right~\cite{wei2022COT,wang2022COT-SC}.

The recently conceptualized Tree of Thoughts (ToT)~\cite{yao2023tree} made significant strides in enabling LLMs to exercise foresight and backtrack for holistic decision-making. Yet, an apparent blind spot has been the oversight of local uncertainties in the intermediate~\cite{si2022prompting}. 
These uncertainties, stemming from LLMs' propensity for varied responses~\cite{semantic_uncertainty}, pose a considerable challenge to the reasoning process. 

One fundamental obstacle is the monumental scale of LLMs, rendering them impervious to fine-tuning.
They predominantly serve as black boxes, with the Bayesian modification to obtain a distribution-based uncertainty qualification~(UQ) being far from practical~\cite{UQ-NLG}. 
However, the LLMs with \textit{emergent abilities} can be perceived as approximately unbiased estimations of our inherently uncertain reality~\cite{Lin2022TeachingMT}. 
This makes them less susceptible to the influence of the out-of-domain data on uncertainty estimation, allowing inference approximation of low training-complexity to work. 
Hence, the complexity caused by vast scale and the fascinating nature associated with it advocate a direct and effective mechanism for dealing with uncertainty.

Our novel solution comes in the form of the \textit{Tree of Uncertain Thoughts} (TouT), a pioneering reasoning framework expressly crafted for LLMs. 
Central to TouT is its ingenious employment of Monte Carlo Dropout~\cite{MC_Dropout} for uncertainty qualification. 
This decision was not arbitrary. 
Given the challenges with LLMs, Monte Carlo Dropout presents an elegant, minimalistic, yet robust technique to gauge uncertainty scores linked with the diverse responses of LLMs at intermediate junctures. 
By integrating this local uncertainty measurement with comprehensive sorting algorithms, TouT bolsters the accuracy of model responses.

To validate our method, we undertook rigorous experimentation on two intricate planning tasks: Game of 24 and Mini Crosswords. 
The experimental results decisively highlight the supremacy of TouT over both ToT and the chain-of-thought prompting techniques.

Our pivotal contributions encapsulate:
\vspace{-0.5em}
\begin{itemize}
    \setlength{\itemsep}{1pt}
	\setlength{\parskip}{0pt}
	\setlength{\parsep}{0pt}
    \item The inception of TouT, a groundbreaking Tree of Uncertain Thoughts framework, ushering in uncertainty-aware inference in LLMs.
    \item The innovative integration of Monte Carlo Dropout for local uncertainty quantification and sorting, amplifying model response confidence.
    \item Thorough experimental validation confirming TouT's dominance over the extant ToT and chain-of-thought prompting standards.
\end{itemize}

%% file: SECTIONS/20_Related_Work/manuscript.tex
\section{Related Work}

\noindent\textbf{Large Language Models.}
The advent and progression of Large Language Models (LLMs)~\cite{Radford2018gpt,Radford2019language,brown2020language,zhang2022opt} have been transformative for the fields of natural language processing and machine learning. 
Central to this transformation is the GPT series, which was spearheaded by Radford \textit{et al.}~\cite{Radford2018gpt}. 
Their seminal work led to the development of subsequent iterations, each building upon the strengths and addressing the challenges of the previous versions. 
While the early GPT models~\cite{Radford2019language,brown2020language} laid the foundation, GPT-4~\cite{openai2023gpt4} marked a paradigm shift in the capabilities and applications of LLMs. 
It showcased enhanced reasoning, comprehension, and generation abilities, bridging gaps previously identified in LLMs. 
Parallelly, models like LLaMA-2~\cite{touvron2023llama} have also contributed significantly to the domain. 
LLaMA-2, in particular, emphasized the intersection of linguistic properties with the deep learning capabilities of LLMs, opening new avenues for research and application.
In this work, our main focus is to propose an innovative thoughts reasoning framework tailored for LLaMA-2, aiming to unravel complex deliberative challenges.

\noindent\textbf{Thoughts Reasoning.}
With the sophistication of LLMs, there arose a need to understand, modulate, and enhance the reasoning capabilities underlying their decision-making processes. 
The initial Input-Output models established a basic framework for how models perceive and respond to given prompts. 
Building on this, the Chain of Thoughts (CoT)~\cite{wei2022COT} model was introduced, emphasizing a chained, sequential approach to decision-making. 
It was further refined with CoT-SC~\cite{wang2022COT-SC}, which provided a more structured and systematic framework for thoughts sequencing. 
More recently, ToT~\cite{yao2023tree} incorporated both a hierarchical and lateral understanding of reasoning, enabling LLMs to not only draw from a depth of knowledge but also to assess and reassess decisions in a tree-like structure. 
This approach facilitated greater foresight, backtracking, and holistic decision-making in LLMs.
Yet, a noticeable gap persisted: the oversight of local uncertainties during intermediate decision-making. 
Addressing this, our work pioneers a framework that synergizes local uncertainty quantification with advanced global algorithms, intending to heighten the accuracy of LLM responses.

\noindent\textbf{Uncertainty Quantification}.  The prediction of uncertainty~\cite{MC_Dropout, BNN} represents the foundation for dependable and consistent automated decision-making, and consequently is receiving increasing attention. 
However, obtaining uncertainty quantification in LLM is very challenging, mainly due to the extremely high dimensionality~\cite{Xiao2022UncertaintyQW}. 
Certain recent methodologies explore the issue of uncertainty quantification with black-box LLMs~\cite{semantic_uncertainty, UQ-NLG}. 
However, these techniques concentrate mostly on free-form question answering. 
Scant research has explored the uncertainty quantification for LLMs in complex reasoning, the emphasis of the present paper.

%% file: SECTIONS/30_Method/manuscript.tex
\section{Method}

Given a deliberate problem with several intermediate steps, our target is to leverage pre-trained large language models for problem-solving and decision-making.
We propose a novel framework with the Tree of Uncertain Thoughts, named TouT, for language model inference,  which mainly consists of two modules, Local Uncertainty Quantification in Section~\ref{sec:luq} and Uncertainty Global Search in Section~\ref{sec:ugs}.

\subsection{Preliminaries}

In this section, we first describe the problem setup and notations and then revisit the tree of thoughts reasoning for LLMs inference.

\noindent\textbf{Problem Setup and Notations.}
Given a pre-trained language model (LM) ($p_\theta$) with parameters $\theta$ and a language sequence $\{x,y,z,s,...\}$, our target is to infer LM ($p_\theta(x)$) for generating answers to a deliberate problem.
For each language sequence $x$ with $t$ tokens, we denote $x = \{x[1], x[2], ..., x[t]\}$.

\noindent\textbf{Revisit Non-Tree-Based Prompting.}
To address the task, Input-output (IO) prompting generated the output $y$ from LM with $x$ as input instructions, which can be denoted as $y\sim p_{\theta}^{\textbf{\texttt{IO}}}(y|x)$.
When it comes to non-trivial questions with multiple steps, Chain-of-thought (CoT) prompting~\cite{wei2022COT} introduced a chain of $n$ thoughts $z_1, z_2, ..., z_n$ to solve the problem, where the output is formulated as $y\sim p_{\theta}^{\textbf{\texttt{COT}}}(y|x, z_1, z_2, ..., z_n)$.
To improve COT further, ensemble-based CoT-SC~\cite{wang2022COT-SC} proposed to generate multiple chains of thoughts and select the highest frequency output.

\noindent\textbf{Revisit Tree-of-Thoughts.}
To solve the problem in a human problem-solving manner, ToT~\cite{yao2023tree} proposed to search over a tree consisting of multiple partial solutions (state $s=\{x, z_{1,2,...,i}\}$) as nodes. 
Given the properties of different problems, ToT first decomposed intermediate thought steps and used a thought generator $G(p_{\theta}, s, k)$ to generate $k$ candidates based on a tree state $s$.
With a set $S$ of different states, they adopted a state evaluator $V(p_{\theta}, S)$ to independently measure the possibility of solving the problem for each state.
Finally, they plugged a search algorithm to select the most promising state for the final output.

However, the ToT reasoning paradigm grapples with the complexities of local uncertainties at intermediate ``thoughts''. 
Given the innate capacity of LLMs to generate a spectrum of responses, these local uncertainties can become significant impediments in the reasoning process. 
We introduce the ``Tree of Uncertain Thoughts" framework to address this challenge, pioneering a shift towards uncertainty-aware inference within LLMs.

\subsection{Local Uncertainty Quantification}\label{sec:luq}

To explicitly quantify the uncertainty for each local response in intermediate steps, we introduce a novel uncertainty evaluator $U(p_{\theta}, S_{1,2,...,m})$ to generate a scalar value to represent the confidence score for each local intermediate states, that is, $U(p_\theta,S)(s)\sim p^{uncertain}(u|s), \forall s\in S$.
Specifically, we are inspired by Monte Carlo Dropout~\cite{MC_Dropout} and generate $S_{1,2,...,m}$ with $m$ sampling steps on LLMs inference.
Meanwhile, we adopt an $ m$-step-based linear interpolation on the input temperature of LLMs to control the quality of responses for each intermediate step. 

After sampling, we compute the variance of values from a set $\{S_{t}^\prime\}$  of $m$ states in this step, where the variance of values is used as the local uncertainty score $u$ for each state.
Such evaluations will enable us to comprehensively evaluate diverse local responses instead of generating candidates using one fixed model temperature. 
Furthermore, we can use quantified states for later global searching to find the correct answers to the problem more confidently.

\input{SECTIONS/30_Method/algo_bfs}

\subsection{Uncertainty-aware Global Search}\label{sec:ugs}

Benefiting from the above uncertainty quantification on local responses, we leverage a novel and explicit uncertainty-aware global search mechanism to select a more precise state.
During searching, we use $v/u$ as the final evaluation score for criteria to finalize the state with the largest score, where $v, u$ denote the value and uncertainty of the state, respectively.
Based on the new criteria, we propose two search algorithms for uncertainty-aware global search.

One is based on Breadth-first search (BFS), TouT-BFS uses a set of the $b$ most confident states per step by selecting from $m$ states using the new score $V_t(S)/U_t(S)$, as illustrated in Algorithm~\ref{alg:tout_bfs}.
The other one is Depth-first search (DFS), we either explore the most confident state in global steps $T$ or use $V(p_\theta,\{s^\prime\})(s)\leq v_{th}$ for a value threshold $v_{th}$ and $U(p_\theta,\{s^\prime\})(s)\geq u_{th}$ for a uncertainty threshold $u_{th}$. 
For both cases, the algorithm backtracks to the parent state of $s$ with the higher value and lower uncertainty and continually finds the correct answers, as shown in Algorithm~\ref{alg:tout_dfs}.

\input{SECTIONS/30_Method/algo_dfs}

%% file: SECTIONS/30_Method/algo_bfs.tex
\begin{algorithm}[t]
\caption{TouT-BFS($x,p_\theta, G, k, V,T,b, U,m$)}\label{alg:tout_bfs}
\begin{algorithmic}
\Require Input $x$, LM $p_\theta$, thought generator $G(\cdot)$, candidates size $k$, states evaluator $V(\cdot)$, global steps $T$, breadth limit $b$, uncertainty evaluator $U(\cdot)$, sampling steps $m$.
\State $S_0 \gets \{x\}$
\For{$t=1,2,..., T$} 
    \State $S_t^\prime \gets \{[s,z] | s\in S_{t-1}, z_t\in G(p_\theta, s, k) \}$
    \State $U_t \gets U(p_\theta, \{S_{t}^\prime\})$
    \State $V_t \gets V(p_\theta, \{S_t^\prime\})$
    \State $S_t \gets \arg\max_{S\subset 
    S_t^\prime, |S|=b}\sum_{s\in S}V_t(S)/U_t(S)$
\EndFor
\State \Return $G(\theta, \arg\max_{s\in S_T} V_t(S)/U_t(S), 1)$
\end{algorithmic}
\end{algorithm}

%% file: SECTIONS/30_Method/algo_dfs.tex
\begin{algorithm}[t]
\caption{TouT-DFS($s,t,p_\theta, G,k V, T, v_{th}, U, u_{th}$)}\label{alg:tout_dfs}
\begin{algorithmic}
\Require Current state $s$, step $t$, LM $p_\theta$, thought generator $G(\cdot)$, candidates size $k$, states evaluator $V(\cdot)$, global steps $T$, state threshold $v_{th}$, uncertainty evaluator $U(\cdot)$, uncertainty threshold $u_{th}$.
\If{$t>T$} \State record output $G(p_\theta, s, 1)$
\EndIf
\For{$s^\prime\in G(p_\theta, s, k)$} 
    \If{$V(p_\theta, \{s^\prime\})(s)>v_{th}$ and $U(p_\theta, \{s^\prime\})(s)<u_{th}$}  
        \State DFS($s^\prime, t+1$)
    \EndIf
\EndFor
\end{algorithmic}
\end{algorithm}

%% file: SECTIONS/40_Experiments/manuscript.tex
\section{Experiments}

\subsection{Experimental setup}

\noindent\textbf{Tasks.}
Game of 24~\cite{yao2023tree} contains 
1,362 games with human solving levels from easy to hard, and a subset of relatively hard games indexed 901-1,000 is used for testing.
The thoughts in this task are decomposed into 3 steps. 
Mini Crosswords~\cite{yao2023tree} includes 156 games of $5 \times 5$ mini crosswords.
For this task, the input is the 5 horizontal clues and 5 vertical clues, and the output should be a board of 25 letters to solve the problem.
This task has 5-10 intermediate steps for solving, such as h1. shown and v5. naled.

\noindent\textbf{Evaluation Metrics.}
For Game of 24, if the output is an equation that uses the input numbers each exactly once equals 24, it is regarded as a success, such as (13-9)*(10-4)=24.
We compute the average success rate of total testing games, where the Breadth-first search algorithm is used for this task.
For Mini Crosswords, we follow ToT~\cite{yao2023tree}, and adopt three levels of success: the accuracy of letters (25 per game), words (10 per game), and games.

\noindent\textbf{Implementation.}
For LLMs, we use officially released LLaMA-2-70B~\cite{touvron2023llama} weights.
Since GPT-4 is more expensive to use, we reproduce all baseline results using the same LLM weight for a fair comparison.
The number of Monte Carlo sampling steps $m$ is 20.
Our experiments are conducted on NVIDIA-A100 GPUs.

\input{SECTIONS/40_Experiments/exp_game}

\input{SECTIONS/40_Experiments/exp_minicross}

\subsection{Comparison to prior work}

In this work, we propose a novel and effective framework for deliberate problem-solving with LLMs inference.
In order to demonstrate the effectiveness of the proposed TouT, we compare it to previous non-tree-based prompting~\cite{wei2022COT,wang2022COT-SC} and tree-of-thoughts~\cite{yao2023tree} methods.

For the Game of 24 task, we report the quantitative comparison results in Table~\ref{tab: exp_game}.
As can be seen, we achieve the best results regarding both $b$=1 and $b$=5 for solving the Game of 24 problem.
In particular, the proposed TouT superiorly outperforms ToT~\cite{yao2023tree}, the current state-of-the-art LLM inference baseline, by 5\% and 9\%.
Furthermore, we achieve significant performance gains compared to previous non-tree-based prompting approaches~\cite{wei2022COT,wang2022COT-SC}.
These significant improvements demonstrate the superiority of our approach in deliberate problem-solving with LLMs inference.

In addition, In addition, significant gains in the Mini Crosswords task can be observed in Table~\ref{tab: exp_minicross}.
Compared to ToT~\cite{yao2023tree}, we achieve the results gains of 2\%, 4\%, and 3\% on letter, word and game. 
We also achieve highly better results against IO and CoT prompting baselines.
These results demonstrate the effectiveness of our approach in using LLMs inference for solving problems.

%% file: SECTIONS/40_Experiments/exp_game.tex
\begin{table}[t]
	\renewcommand\tabcolsep{6.0pt}
	\centering
	\scalebox{0.95}{
		\begin{tabular}{lc}
			\toprule
   Method & Success Rate (\%) \\
   \midrule
   IO prompt & 6 \\
   CoT prompt~\cite{wei2022COT} & 3 \\
   CoT-SC~\cite{wang2022COT-SC} & 7.2 \\ \hline
   ToT~\cite{yao2023tree} ($b$=1) & 37 \\
   TouT (ours, $b$=1) & \bf 42\\ \hline
   ToT~\cite{yao2023tree} ($b$=5) & 56 \\
   TouT (ours, $b$=5) & \bf 65\\
   \bottomrule
			\end{tabular}}
   \caption{Quantitative results of Game of 24.}
   \label{tab: exp_game}
\end{table}

%% file: SECTIONS/40_Experiments/exp_minicross.tex
\begin{table}[t]
	\renewcommand\tabcolsep{6.0pt}
	\centering
	\scalebox{0.95}{
		\begin{tabular}{lccc}
			\toprule
   \multirow{2}{*}{Method} & \multicolumn{3}{c}{Success Rate (\%)} \\
   & Letter & Word & Game \\ 
   \midrule
   IO prompt & 29.5 & 10 & 0 \\
   CoT prompt~\cite{wei2022COT} & 33.2 & 10.8 & 0 \\
   ToT~\cite{yao2023tree} & 59 & 48 & 12 \\
   TouT (ours) & \bf 61 & \bf 52	& \bf 15 \\
   ToT~\cite{yao2023tree} + best state & 62.2	& 53.9	& 21 \\		
   TouT (ours) + best state & \bf 64.5 & \bf 58.2 & \bf 29 \\
   \bottomrule
			\end{tabular}}
   \caption{Quantitative results of Mini Crosswords.}
   \label{tab: exp_minicross}
\end{table}

%% file: SECTIONS/50_Ablation_Study/manuscript.tex
\input{SECTIONS/50_Ablation_Study/ab_component}

\subsection{Experimental analysis}

In this section, we performed ablation studies to demonstrate the benefit of introducing the Local Uncertainty Quantification and Uncertainty-aware Global Search modules.
We also conducted extensive experiments to explore the impact of Monte Carlo sampling steps in uncertainty quantification.

\noindent{\textbf{Local Uncertainty Quantification \& Uncertainty-aware Global Search.}
In order to demonstrate the effectiveness of the introduced Local Uncertainty Quantification (LUQ) and Uncertainty-aware Global Search (UGS), we ablate the necessity of each module and report the quantitative results in Table~\ref{tab: exp_ablation}.
As can be observed, adding LUQ to the vanilla baseline highly increases the results of 4\%, 2\%, 3.6\%, and 6\%, which validates the benefit of LUQ in quantifying the uncertainty for each local response in intermediate steps.
Meanwhile, introducing only UGS in the baseline increases the performance regarding all metrics.
More importantly, incorporating LUQ and UGS into the baseline significantly raises the performance, and achieves the best.
These improving results validate the importance of local uncertainty quantification and uncertainty-aware global search in deliberate problem-solving with LLMs inference.

\input{SECTIONS/50_Ablation_Study/ab_steps}

\noindent{\textbf{Impact of Monte Carlo sampling steps.}}
The number of Monte Carlo sampling steps used in the proposed LUQ affects the selected state in global searching for the final answer.
To explore such effects more comprehensively, we varied the number of sampling steps from $\{5,10,20,50,100\}$.
We report the comparison results of Game of 24 and Mini Crosswords in Table~\ref{tab: exp_steps}.
When the number of Monte Carlo sampling steps is 20, we achieve the best performance regarding all metrics.
With increased depth from 5 to 20, the proposed TouT consistently increases performance as best states are extracted from more quantified local responses.
Nevertheless, increasing the steps from 20 to 100 will not continually improve the results since 20 steps might be enough to extract the best state from these quantified states for addressing our deliberate problems with at most 10 intermediate thoughts.

%% file: SECTIONS/50_Ablation_Study/ab_component.tex
\begin{table}[t]
	\renewcommand\tabcolsep{6.0pt}
	\centering
	\scalebox{0.95}{
		\begin{tabular}{cccccc}
			\toprule
   \multirow{2}{*}{LUQ} & \multirow{2}{*}{UGS} & \multirow{2}{*}{Game of 24} & \multicolumn{3}{c}{Mini Crosswords} \\
   & & & Letter & Word & Game \\ 
   \midrule
   \xmark & \xmark & 56 & 61 & 52 & 15 \\
    \cmark & \xmark & 60 & 62 & 55.6 & 21 \\
\xmark & \cmark & 61 & 63 & 56.1 & 23 \\
      \cmark & \cmark & \bf 65 & \bf 64.5	& \bf 58.2 & \bf 29 \\
   \bottomrule
			\end{tabular}}
   \caption{Ablation study on Local Uncertainty Quantification (LUQ) and Uncertainty-aware Global Search (UGS).}
   \label{tab: exp_ablation}
\end{table}

%% file: SECTIONS/50_Ablation_Study/ab_steps.tex
\begin{table}[t]
	\renewcommand\tabcolsep{6.0pt}
	\centering
	\scalebox{0.95}{
		\begin{tabular}{lcccc}
			\toprule
   \multirow{2}{*}{Steps} & \multirow{2}{*}{Game of 24} & \multicolumn{3}{c}{Mini Crosswords} \\
   & & Letter & Word & Game \\ 
   \midrule
   5 & 61 & 62.7 & 55.3 & 22 \\
   10 & 63 & 63.2 & 56.8 & 25 \\
   20 & \bf 65 & \bf 64.5 & \bf 58.2 & \bf 29 \\
   50 & 65 & 64.2 & 58.1 & 29 \\
   100 & 64 & 63.9 & 57.8 & 28 \\
   \bottomrule
			\end{tabular}}
   \caption{Exploration studies on the Monte Carlo sampling steps in Local Uncertainty Quantification.}
   \label{tab: exp_steps}
\end{table}

%% file: SECTIONS/60_Conclusion/manuscript.tex
\section{Conclusion}

In this work, we present TouT, a novel framework with the Tree of Uncertain Thoughts for large-scale language model inference.
We leverage Monte Carlo Dropout for local uncertainty quantification on diverse responses at intermediate steps.
Furthermore, we integrate local uncertainty into global sorting to amplify model response confidence. 
Experimental results on Game of 24 and Mini Crosswords comprehensively demonstrate the state-of-the-art superiority against previous ToT and CoT prompting methods.
Extensive ablation studies also validate the importance of local uncertainty quantification and Local Uncertainty Sorting in generating more accurate answers for LLMs inference to solve deliberate problems.